\documentclass[letterpaper]{article} 
\usepackage{aaai25}  
\usepackage{times}

\usepackage{latexsym}
\usepackage{amssymb}
\usepackage{amsmath}
\usepackage{amsthm}
\usepackage{booktabs}
\usepackage{enumitem}
\usepackage{graphicx}
\usepackage{array}
\usepackage{multirow}
\usepackage{amsfonts}
\usepackage{subcaption}
\usepackage[ruled,vlined,linesnumbered]{algorithm2e}
\usepackage{xspace}
\usepackage{caption}
\usepackage{makecell}
\usepackage[table]{xcolor}
\usepackage{natbib}
\usepackage{tikz}

\newcommand{\fluents}{\ensuremath{F}\xspace}
\newcommand{\actions}{\ensuremath{A}\xspace}
\newcommand{\init}{\ensuremath{I}\xspace}
\newcommand{\goal}{\ensuremath{G}\xspace}
\newcommand{\cost}{\ensuremath{c}\xspace}

\newcommand{\task}{\ensuremath{{\cal P}}\xspace}
\newcommand{\plan}{\ensuremath{\pi}\xspace}
\newcommand{\plans}{\ensuremath{\Pi}\xspace}
\newcommand{\stripstask}{\ensuremath{\task=\langle \fluents, \actions, \init, \goal \rangle}\xspace}
\newcommand{\state}{\ensuremath{s}\xspace}

\newcommand{\action}{\ensuremath{a}\xspace}
\newcommand{\name}{\ensuremath{\textsc{name}}\xspace}
\newcommand{\precondition}{\ensuremath{\textsc{pre}}\xspace}

\newcommand{\addeffects}{\ensuremath{\textsc{add}}\xspace}
\newcommand{\deleffects}{\ensuremath{\textsc{del}}\xspace}
\newcommand{\actionapplication}{\ensuremath{\gamma}\xspace}
\newcommand{\planapplication}{\ensuremath{\Gamma}\xspace}
\newcommand{\alternativeplans}[1]{\ensuremath{\plans^{#1}}\xspace}

\newtheorem{definition}{Definition}

\title{Counterfactual Reasoning in Automated Planning}
\author{
    Alberto Pozanco,
    Daniel Borrajo,
    Manuela Veloso\footnote{Work completed while at J.P. Morgan AI Research.}
}
\affiliations{
    J.P. Morgan AI Research \\
    \{alberto.pozancolancho\}@jpmorgan.com, \{name.surname\}@jpmorgan.com
}

\begin{document}

\maketitle

\begin{abstract}
Automated planning traditionally assumes that all aspects of a planning task—initial state, goals, and available actions—are fully specified in advance, an approach well-suited to domains with fixed rules and deterministic execution.
However, real-world planning often requires flexibility, allowing for deviations from the original task parameters in response to unforeseen circumstances or to improve outcomes. 
This paper surveys existing works on counterfactual reasoning in automated planning, categorizing them by what elements are changed, when the reasoning is triggered, and why and how these changes are made. 
We conclude by discussing key findings and outlining open research questions to guide future work in this area.
\end{abstract}

\section{Introduction}

Automated planning involves determining a sequence of actions, known as a plan, to achieve specific goals from an initial state~\cite{DBLP:books/daglib/0014222}. 
Traditionally, most research in automated planning assumes that planning tasks are provided as input, with the initial state, desired goals, and available actions fully specified in advance. 
This approach is well-suited for applications like games, where rules are clearly defined and static. 
However, in many real-world planning scenarios, it is often acceptable to deviate slightly from the predefined task parameters, allowing for more flexibility and adaptability.

For instance, in real-world scenarios, it may be beneficial to relax the original goals, such as parking a car further from the intended destination when encountering a roadblock, or allowing a robot to switch to a fast-charging station if one is discovered during plan execution. 
These examples illustrate how goals can be adjusted during execution due to unforeseen circumstances. Additionally, there is value in considering changes to the input task during the planning phase. 
For example, traffic authorities might evaluate the potential benefits of constructing a new road to alleviate city traffic congestion. 
Similarly, regulatory bodies could explore policy adjustments to encourage or discourage specific behaviors. 
Web developers might also reconsider the available actions within their applications to enhance user experience by simplifying interactions.

Although some research has explored the use of counterfactuals to explain solutions to planning tasks~\cite{krarup2024explaining}, there is currently no unified framework or terminology to describe approaches that allow for changes or deviations from the original input planning task.
In this position paper, we survey various works that employ counterfactual reasoning~\cite{pearl2013structural} within the context of planning tasks. 
We categorize these works based on the following four criteria:
\begin{itemize}
    \item \textbf{What}: This criterion examines the specific elements that are altered during counterfactual reasoning. Changes may involve various aspects of the planning task, such as modifying the available actions, adjusting the initial state, or redefining the goals to be achieved.

    \item \textbf{Why}: This criterion explores the underlying objectives for initiating counterfactual reasoning. It may be employed to transform an unsolvable task into a solvable one, to allow for better plans, or to impose constraints on agent behavior within the environment, among other objectives.

    \item \textbf{When}: This refers to the time at which the counterfactual reasoning is triggered. It can occur either offline, prior to the planning process, or during the execution of the plan.

    \item \textbf{How}: This refers to the methodology used to perform counterfactual reasoning.
\end{itemize}
We conclude the paper by summarizing the key findings and highlighting several open research questions within the domain of counterfactual reasoning in automated planning. 
These questions aim to guide future exploration and development in this evolving field.
\section{Planning Formalism}

We formally define a planning task as follows:

\begin{definition}\label{def:strips-plan-task}
  A {\sc strips} \textbf{planning task}
is a tuple \stripstask, where \fluents is a set of fluents, \actions is a set of
 actions, $\init \subseteq \fluents$ is an initial state, and $\goal\subseteq \fluents$ is a goal specification.  
\end{definition}

A state $\state \subseteq \fluents$ is a set of fluents that are true at a given time.
A state $\state \subseteq \fluents$ is a goal state iff $\goal \subseteq \state$.
Each action $\action \in \actions$ is described by its name $\name(\action)$, a set of positive and negative preconditions $\precondition^+(\action)$ and $\precondition^-(\action)$, add effects $\addeffects(\action)$, delete effects $\deleffects(\action)$, and cost $\cost(\action)$.
An action \action is applicable in a state \state iff $\precondition^+(\action)\subseteq~s$ and $\precondition^-(\action) \cap s = \emptyset$.
We define the result of applying an action in a state as $\actionapplication(\state,\action)=(\state \setminus \deleffects(\action)) \cup \addeffects(\action)$.
We assume $\deleffects(\action) \cap \addeffects(\action) = \emptyset$.
A sequence of actions $\plan=(\action_1,\ldots,\action_n)$ is applicable in a state $\state_0$ if there are states $(\state_1.\ldots,\state_n)$ such that $\action_i$ is applicable in $\state_{i-1}$ and $\state_i=\actionapplication(\state_{i-1},\action_i)$.
The resulting state after applying a sequence of actions is $\planapplication(\state,\pi)=\state_n$, and $\cost(\plan) = \sum_{i}^n \cost(\action_i)$ denotes the cost of $\plan$.
A state $\state$ is reachable from state $\state^\prime$ iff there exists an applicable action sequence \plan such that $s \subseteq \planapplication(\state^\prime,\pi)$.
The solution to a planning task $\task$ is a plan, i.e., a sequence of actions $\plan$ such that $\goal \subseteq \planapplication(\init,\pi)$.
We denote as $\Pi(\task)$ the set of all solution plans to planning task $\task$. 
Also, given a plan \plan, we denote its alternatives, i.e., all the other sequence of actions that can solve \task as $\alternativeplans{\plan} = \Pi(\task) \setminus \pi$.
A plan with minimal cost is optimal.

\section{Counterfactual Reasoning in Automated Planning}

\begin{figure*}
    \centering
    \includegraphics[width=0.45\linewidth]{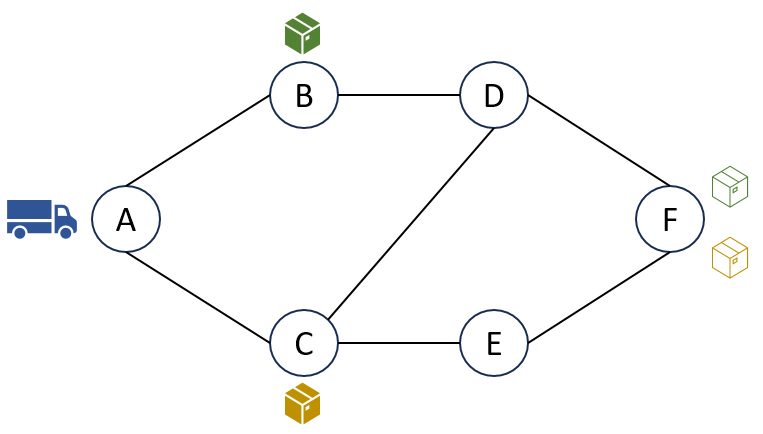}
    \caption{\textsc{logistics} task where a truck must deliver two packages by moving them from their current (filled) locations to their goal (empty) destinations.}
    \label{fig:placeholder}
\end{figure*}

In this section, we introduce and categorize the literature on what we term \emph{counterfactual reasoning}—that is, modifying the input planning task to achieve different outcomes. 
The subsections will be organized primarily by the elements of the planning task that are modified (the "what"). 
Within each subsection, we will further discuss the underlying objectives for these modifications (the "why"), the timing at which counterfactual reasoning is triggered (the "when"), and the methodologies used to perform counterfactual reasoning (the "how").
Throughout the section, we use the \textsc{logistics} task depicted in Figure~\ref{fig:placeholder} in which a truck must deliver two packages by moving them through the network as a running example.

\subsection{Changing $F$}

The set of fluents in a planning task defines the language universe—that is, all entities and symbols utilized throughout the planning process.
While some works attempt to learn these symbols from various inputs~\cite{asai2018classical,bonet2020learning}, the set of fluents is seldom treated as a modifiable component.
A notable recent exception is the work of \citeauthor{correa2024planning} (\citeyear{correa2024planning}), which investigates planning with object creation.
By enabling the creation and removal of objects, this approach effectively allows for dynamic modification of the set of fluents $F$.
Theoretically, the authors demonstrate that planning with object creation is semi-decidable, implying that, in general, no algorithm can determine whether a given task is unsolvable.
Practically, they introduce a PDDL~\cite{haslum2019introduction} extension that allows for object creation/removal in the effects of actions.
This extension facilitates more natural problem formulations, permitting objects to be created or discarded as needed, rather than requiring all potential objects to be specified in advance.
This approach has significant implications beyond enhancing search efficiency by reducing state size and minimizing unnecessary objects. 

It also facilitates counterfactual reasoning, enabling researchers to hypothesize offline about the types of plans that might emerge if the number of objects were different.
For example, in the \textsc{logistics} task illustrated in Figure~\ref{fig:placeholder}, the presence of an additional truck in the grid would allow for more cost-effective plans, as each truck could be assigned to deliver individual packages. 
This would reduce the need for a single truck to make multiple trips back and forth.

\subsection{Changing $A$}
The set of actions in a planning task describes the dynamics of the world and how agents operating in the environment can influence it.
Similar to our treatment of the set of fluents, we exclude works that concentrate on model learning or acquisition—specifically, those that involve generating the set of actions for a planning task~\cite{arora2018review,aineto2019learning}. 
Instead, we focus on studies that modify an existing or predefined set of actions $A$.
This process is commonly referred to in the literature as model repair~\cite{bercher2025survey}. 
However, the terminology itself suggests a conceptual implication: model repair presupposes that the set of actions is flawed or incorrect and requires rectification.
These rectifications have different objectives.

Most works have focused on how to change the set of actions such that a planning task becomes solvable.
\citeauthor{gragera2023planning} (\citeyear{gragera2023planning}) do not assume any input knowledge and create a planning compilation that simultaneously modifies the existing actions and solves the planning task.
A modified version of this compilation~\cite{gragera2025gains} can also accommodate knowledge in the form of partial observations or plans that must be satisfied in the resulting model.
Other works, such as~\cite{lin2023towards,lin2025told}, assume that  plans—required to be valid or invalid in the resulting model—are provided as input, and employ a hitting-set based approach to modify the existing actions.
However, none of these works address the generation of new actions from scratch; they focus solely on modifying the existing actions.

Other works aim to change the original actions such that a given plan is optimal in the resulting model.
This concept is commonly referred to in the literature as Model Reconciliation~\cite{chakraborti2017plan}.
Some approaches do it by changing the preconditions and effects of the actions~\cite{sreedharan2021foundations}, while others focus on modifying the actions costs~\cite{morales2024learning}.

Finally, some works focus on environment redesign~\cite{zhang2009general}, which is the task of modifying a set of actions such that the resulting model allows or disallows solutions of a desired property.
Some examples include goal recognition~\cite{keren2021goal} or goal privacy~\cite{pozanco2024generalising} design, where the aim is to facilitate the identification/obfuscation of the goal of an agent acting in the environment.
These approaches explore the space of permissible action removals, generating a new set of actions $A^\prime$, and then evaluate a specific metric in the modified model.
The search process concludes when this metric is optimized, resulting in an environment that restricts agent behavior to achieve the intended objective.
At the extreme end of this spectrum, \cite{pozanco2026planning} modify the set of actions so that no plan exists to solve the planning task, thereby completely limiting agent behavior.

We identify two interconnected gaps in the literature.
First, existing research primarily focuses on either completely removing actions or modifying the preconditions or effects of current actions. 
However, these works do not address the generation of entirely new actions from scratch.
Second, none of these studies consider modifying actions with the explicit goal of enabling less costly plans.
While this objective has been explored in related fields, such as Markov Decision Processes—specifically under the concept of Configurable MDPs~\cite{silva2019theoretical, metelli2018configurable}—it has not received attention within the planning community.

Both capabilities would be highly beneficial in the context of the \textsc{logistics} running example.
For instance, introducing new actions, such as additional routes directly connecting $\mathsf{B}$ or $\mathsf{C}$ to $\mathsf{F}$, would allow the truck to deliver packages more efficiently.
This form of counterfactual reasoning can help identify minimal changes to the planning model that facilitate the existence of more cost-effective plans, which may be applicable in practical scenarios.
Reasoning about the existence of new or modified actions can be both an offline process where we consider the model as a whole, or an online process that is triggered only upon reaching some pre-defined states during plan execution. 
For example, it may be triggered during replanning if the new plan is significantly more costly than the original plan whose execution failed.

\subsection{Changing $I$}
The initial state of a planning task describes the current state of the world.
There are two main motivations identified in the literature for changing the initial state. 

Similar to the case of actions, some works reason about modifying the initial state with the goal of making an unsolvable planning task solvable, aiming to generate explanations for the unsolvability~\cite{gobelbecker2010coming,sreedharan2019can}.
This initial state may refer either to the state at planning time, before any actions are executed, or to the current state after replanning has been triggered.

On the other hand, other works reason about modifying the initial state aiming to improve the quality of the generated plans.
This is the case of the works on planning opportunities~\cite{borrajo2021computing,borrajo2021intelligent}, which analyze offline which would be the set of (static) facts that if took a different truth value during execution, would enable the existence of better plans.
These facts are then monitored at execution time in order to trigger opportunistic replanning in the pursuit of better plans.
This could be illustrated by a hypothetical truck located in an isolated location $\mathsf{X}$ in our running example.
These approaches would identify the value in monitoring any potential connection between $\mathsf{X}$ and the rest of the network, as such an event would present an opportunity to enable more cost-effective plans.

In all these cases, the counterfactual reasoning is restricted to propositions that are true in the initial state. 
It would be interesting to build on recent work in object creation~\cite{correa2024planning} to generate and systematically explore new initial state configurations.

\subsection{Changing $G$}
The goal specification of a planning task defines the subset of properties that must hold in any goal state we aim to achieve with our plan.

Works on Goal-Driven Autonomy~\cite{molineaux2010goal,jaidee2011integrated} focus on the lifecycle of goals, generating and modifying goals based on their potential emergence~\cite{burns2012anticipatory,fuentetaja2018anticipation,pozanco2018learning,pozanco2024computing} or in response to the behavior of other agents~\cite{pozanco2018counterplanning}.
However, none of these studies explicitly perform counterfactual reasoning in which multiple alternative goals are evaluated.

Some works capture the flexibility of changing goals by modeling the problem as an oversubscription problem~\cite{smith2004choosing}, where multiple goal conditions are associated with different costs~\cite{cushing2008replanning}. 
In these approaches, the planner determines at planning time the best set of goals to achieve under the given conditions. The main limitation of these methods is that the allowed goal conditions must be explicitly specified in advance. Other research challenges this assumption by enabling goal changes during plan execution without requiring prior specification of potential goal candidates~\cite{pozanco2023generating}. 
In these cases, the problem of generating a new goal is formulated as an optimization problem that considers (i) the distance between the original goal and the candidate goal, (ii) the consistency of the candidate goal with respect to the actions already executed, and (iii) the cost of the plan required to achieve the new candidate goal. 
By adjusting the weights of these three objectives, it is possible to generate various counterfactual goals and select the most suitable one for a given scenario.

Returning to our running example, the truck could engage in counterfactual reasoning if, due to unforeseen circumstances, it consumes more gas than anticipated while executing the original plan. 
At this point, the agent might consider alternative goals, such as delivering only one of the packages or unloading both packages at location $\mathsf{D}$. 
This latter option would represent a reasonable compromise between the original destinations and the truck’s current gas level.

\section{Discussion}
In this position paper, we have motivated the need for counterfactual reasoning in planning by relaxing the common assumption that all parameters of the input task are fixed. 
This capability is intuitive in many real-world planning applications, where practitioners may prefer or require the planning architecture to consider deviations from the initial scenario to enhance flexibility and adaptability. 
We have also reviewed works that incorporate counterfactual reasoning at various levels, categorizing them according to which elements of the planning task are modified, when the reasoning is initiated, and the motivations and methods behind these changes.

The survey reveals that counterfactual reasoning in automated planning is a rapidly evolving field, yet several significant gaps remain that limit its practical impact and realism. Most notably, there are very few architectures that combine even two of the major counterfactual mechanisms discussed—such as action modification, goal adaptation, or initial state changes—within a single, unified framework. Existing approaches tend to treat these elements in isolation, which restricts the ability to model the complex, interdependent decisions encountered in real-world scenarios. Developing integrated architectures that can simultaneously reason about and modify multiple aspects of the planning task would result in more realistic and robust systems.

A related trend is the increasing recognition of the need for planners to make dynamic decisions about when and how to apply counterfactual reasoning. In oversubscription planning~\cite{smith2004choosing}, for example, the planner autonomously selects which goals to pursue based on contextual costs and constraints. Extending this paradigm, future research should focus on embedding the decision to generate new actions or change goals directly into the planner’s reasoning process. Rather than relying on static, pre-specified alternatives, planners could evaluate, in real time, whether introducing new actions or adapting goals would yield better outcomes given the current state and objectives. This would allow for more flexible and adaptive planning, especially in environments where unforeseen opportunities or obstacles frequently arise.

Despite progress in areas such as model repair~\cite{bercher2025survey}, environment redesign~\cite{keren2021goal,pozanco2024generalising}, and goal-driven autonomy~\cite{molineaux2010goal}, the literature still lacks systematic approaches for evaluating multiple counterfactual alternatives. Most current methods do not provide mechanisms for planning architectures to compare the relative benefits of different possible changes—such as generating new actions versus modifying goals—or to select the most advantageous option for a given scenario. Embedding such comparative reasoning within planning architectures would enable the generation and assessment of alternative scenarios, leading to more informed and effective decision-making.

In summary, recent work in counterfactual reasoning is making planning systems more flexible and adaptable. Still, there is a clear need for integrated architectures and built-in decision-making processes. Progress in this area will depend on developing systems that combine different counterfactual techniques and give planners and planning architectures the ability to decide when and how to use them. Tackling these challenges will help make planning more realistic and context-aware, especially in complex environments. These directions offer promising opportunities for future research and could greatly improve the effectiveness of automated planning systems.
\section*{Disclaimer}

This paper was prepared for informational purposes by the Artificial Intelligence Research group of JPMorgan Chase \& Co. and its affiliates ("JP Morgan'') and is not a product of the Research Department of JP Morgan. JP Morgan makes no representation and warranty whatsoever and disclaims all liability, for the completeness, accuracy or reliability of the information contained herein. This document is not intended as investment research or investment advice, or a recommendation, offer or solicitation for the purchase or sale of any security, financial instrument, financial product or service, or to be used in any way for evaluating the merits of participating in any transaction, and shall not constitute a solicitation under any jurisdiction or to any person, if such solicitation under such jurisdiction or to such person would be unlawful.
\noindent © 2026 JPMorgan Chase \& Co. All rights reserved.

\bibliography{aaai25}

@inproceedings{gobelbecker2010coming,
  title={Coming up with good excuses: What to do when no plan can be found},
  author={G{\"o}belbecker, Moritz and Keller, Thomas and Eyerich, Patrick and Brenner, Michael and Nebel, Bernhard},
  booktitle={Proceedings of the international conference on automated planning and scheduling},
  volume={20},
  pages={81--88},
  year={2010}
}

@inproceedings{sreedharan2019can,
  title={Why can't you do that hal? explaining unsolvability of planning tasks},
  author={Sreedharan, Sarath and Srivastava, Siddharth and Smith, David and Kambhampati, Subbarao},
  booktitle={International Joint Conference on Artificial Intelligence},
  year={2019}
}

@inproceedings{borrajo2021computing,
  title={Computing opportunities to augment plans for novel replanning during execution},
  author={Borrajo, Daniel and Veloso, Manuela},
  booktitle={Proceedings of the International Conference on Automated Planning and Scheduling},
  volume={31},
  pages={51--55},
  year={2021}
}

@inproceedings{borrajo2021intelligent,
  title={Intelligent Execution through Plan Analysis},
  author={Borrajo, Daniel and Veloso, Manuela},
  booktitle={2021 IEEE/RSJ International Conference on Intelligent Robots and Systems (IROS)},
  pages={3162--3167},
  year={2021},
  organization={IEEE}
}

@incollection{pozanco2023generating,
  title={Generating Replanning Goals Through Multi-Objective Optimization in Response to Execution Observation},
  author={Pozanco, Alberto and Borrajo, Daniel and Veloso, Manuela},
  booktitle={ECAI 2023},
  pages={1898--1905},
  year={2023},
  publisher={IOS Press}
}

@inproceedings{pozanco2024computing,
  title={Computing planning centroids and minimum covering states using symbolic bidirectional search},
  author={Pozanco, Alberto and Torralba, {\'A}lvaro and Borrajo, Daniel},
  booktitle={Proceedings of the International Conference on Automated Planning and Scheduling},
  volume={34},
  pages={455--463},
  year={2024}
}

@inproceedings{burns2012anticipatory,
  title={Anticipatory on-line planning},
  author={Burns, Ethan and Benton, J and Ruml, Wheeler and Yoon, Sungwook and Do, Minh},
  booktitle={Proceedings of the International Conference on Automated Planning and Scheduling},
  volume={22},
  pages={333--337},
  year={2012}
}

@article{fuentetaja2018anticipation,
  title={Anticipation of goals in automated planning},
  author={Fuentetaja, Raquel and Borrajo, Daniel and de la Rosa, Tom{\'a}s},
  journal={AI Communications},
  volume={31},
  number={2},
  pages={117--135},
  year={2018},
  publisher={SAGE Publications Sage UK: London, England}
}

@article{pozanco2018learning,
  title={Learning-driven goal generation},
  author={Pozanco, Alberto and Fern{\'a}ndez, Susana and Borrajo, Daniel},
  journal={AI Communications},
  volume={31},
  number={2},
  pages={137--150},
  year={2018},
  publisher={SAGE Publications Sage UK: London, England}
}

@inproceedings{molineaux2010goal,
  title={Goal-driven autonomy in a Navy strategy simulation},
  author={Molineaux, Matthew and Klenk, Matthew and Aha, David},
  booktitle={Proceedings of the AAAI Conference on Artificial Intelligence},
  volume={24},
  number={1},
  pages={1548--1554},
  year={2010}
}

@inproceedings{jaidee2011integrated,
  title={Integrated learning for goal-driven autonomy},
  author={Jaidee, Ulit and Munoz-Avila, H{\'e}ctor and Aha, David W},
  booktitle={IJCAI Proceedings-International Joint Conference on Artificial Intelligence},
  volume={22},
  number={3},
  pages={2450},
  year={2011}
}

@inproceedings{correa2024planning,
  title={Planning with object creation},
  author={Corr{\^e}a, Augusto B and De Giacomo, Giuseppe and Helmert, Malte and Rubin, Sasha},
  booktitle={Proceedings of the International Conference on Automated Planning and Scheduling},
  volume={34},
  pages={104--113},
  year={2024}
}

@article{aineto2019learning,
  title={Learning action models with minimal observability},
  author={Aineto, Diego and Celorrio, Sergio Jim{\'e}nez and Onaindia, Eva},
  journal={Artificial Intelligence},
  volume={275},
  pages={104--137},
  year={2019},
  publisher={Elsevier}
}

@inproceedings{gragera2023planning,
  title={A planning approach to repair domains with incomplete action effects},
  author={Gragera, Alba and Fuentetaja, Raquel and Garc{\'\i}a-Olaya, {\'A}ngel and Fern{\'a}ndez, Fernando},
  booktitle={Proceedings of the International Conference on Automated Planning and Scheduling},
  volume={33},
  pages={153--161},
  year={2023}
}

@inproceedings{pozanco2024generalising,
  title={Generalising planning environment redesign},
  author={Pozanco, Alberto and Pereira, Ramon Fraga and Borrajo, Daniel},
  booktitle={Proceedings of the AAAI Conference on Artificial Intelligence},
  volume={38},
  number={18},
  pages={20230--20237},
  year={2024}
}

@inproceedings{keren2021goal,
  title={Goal recognition design-survey},
  author={Keren, Sarah and Gal, Avigdor and Karpas, Erez},
  booktitle={Proceedings of the Twenty-Ninth International Conference on International Joint Conferences on Artificial Intelligence},
  pages={4847--4853},
  year={2021}
}

@inproceedings{bercher2025survey,
  title={A Survey on Model Repair in AI Planning},
  author={Bercher, Pascal and Sreedharan, Sarath and Vallati, Mauro},
  booktitle={34th International Joint Conference on Artificial Intelligence},
  year={2025},
  organization={IJCAI Organization}
}

@inproceedings{lin2025told,
  title={Told You That Will Not Work: Optimal Corrections to Planning Domains Using Counter-Example Plans},
  author={Lin, Songtuan and Grastien, Alban and Shome, Rahul and Bercher, Pascal},
  booktitle={Proceedings of the AAAI Conference on Artificial Intelligence},
  volume={39},
  number={25},
  pages={26596--26604},
  year={2025}
}

@inproceedings{lin2023towards,
  title={Towards automated modeling assistance: An efficient approach for repairing flawed planning domains},
  author={Lin, Songtuan and Grastien, Alban and Bercher, Pascal},
  booktitle={Proceedings of the AAAI Conference on Artificial Intelligence},
  volume={37},
  number={10},
  pages={12022--12031},
  year={2023}
}

@article{cushing2008replanning,
  title={Replanning as a Deliberative Re-selection of Objectives},
  author={Cushing, William and Benton, J and Kambhampati, Subbarao},
  journal={Arizona State University CSE Department TR},
  year={2008}
}

@inproceedings{silva2019theoretical,
  title={A theoretical and algorithmic analysis of configurable MDPs},
  author={Silva, Rui and Farina, Gabriele and Melo, Francisco S and Veloso, Manuela},
  booktitle={Proceedings of the International Conference on Automated Planning and Scheduling},
  volume={29},
  pages={455--463},
  year={2019}
}

@inproceedings{smith2004choosing,
  title={Choosing Objectives in Over-Subscription Planning.},
  author={Smith, David E},
  booktitle={ICAPS},
  volume={4},
  pages={393},
  year={2004}
}

@book{DBLP:books/daglib/0014222,
  author       = {Malik Ghallab and
                  Dana S. Nau and
                  Paolo Traverso},
  title        = {Automated planning - theory and practice},
  publisher    = {Elsevier},
  year         = {2004},
  isbn         = {978-1-55860-856-6},
  bibsource    = {dblp computer science bibliography, https://dblp.org}
}

@inproceedings{krarup2024explaining,
  title={Explaining plan quality differences},
  author={Krarup, Benjamin and Coles, Amanda and Long, Derek and Smith, David E},
  booktitle={Proceedings of the International Conference on Automated Planning and Scheduling},
  volume={34},
  pages={324--332},
  year={2024}
}

@inproceedings{asai2018classical,
  title={Classical planning in deep latent space: Bridging the subsymbolic-symbolic boundary},
  author={Asai, Masataro and Fukunaga, Alex},
  booktitle={Proceedings of the aaai conference on artificial intelligence},
  volume={32},
  number={1},
  year={2018}
}

@incollection{bonet2020learning,
  title={Learning First-Order Symbolic Representations for Planning from the Structure of the State Space},
  author={Bonet, Blai and Geffner, Hector},
  booktitle={ECAI 2020},
  pages={2322--2329},
  year={2020},
  publisher={IOS Press}
}

@article{arora2018review,
  title={A review of learning planning action models},
  author={Arora, Ankuj and Fiorino, Humbert and Pellier, Damien and M{\'e}tivier, Marc and Pesty, Sylvie},
  journal={The Knowledge Engineering Review},
  volume={33},
  pages={e20},
  year={2018},
  publisher={Cambridge University Press}
}

@article{sreedharan2021foundations,
  title={Foundations of explanations as model reconciliation},
  author={Sreedharan, Sarath and Chakraborti, Tathagata and Kambhampati, Subbarao},
  journal={Artificial Intelligence},
  volume={301},
  pages={103558},
  year={2021},
  publisher={Elsevier}
}

@inproceedings{metelli2018configurable,
  title={Configurable Markov decision processes},
  author={Metelli, Alberto Maria and Mutti, Mirco and Restelli, Marcello},
  booktitle={International Conference on Machine Learning},
  pages={3491--3500},
  year={2018},
  organization={PMLR}
}

@article{zhang2009general,
  title={A general approach to environment design with one agent},
  author={Zhang, Haoqi and Chen, Yiling and Parkes, David},
  year={2009},
  publisher={Morgan Kaufmann Publishers Inc.}
}

@article{pearl2013structural,
  title={Structural counterfactuals: A brief introduction},
  author={Pearl, Judea},
  journal={Cognitive science},
  volume={37},
  number={6},
  pages={977--985},
  year={2013},
  publisher={Wiley Online Library}
}

@book{haslum2019introduction,
  title={An introduction to the planning domain definition language},
  author={Haslum, Patrik and Lipovetzky, Nir and Magazzeni, Daniele and Muise, Christian and Brachman, Ronald and Rossi, Francesca and Stone, Peter},
  volume={13},
  year={2019},
  publisher={Springer}
}

@article{chakraborti2017plan,
  title={Plan explanations as model reconciliation: Moving beyond explanation as soliloquy},
  author={Chakraborti, Tathagata and Sreedharan, Sarath and Zhang, Yu and Kambhampati, Subbarao},
  journal={arXiv preprint arXiv:1701.08317},
  year={2017}
}

@inproceedings{pozanco2018counterplanning,
  title={Counterplanning using Goal Recognition and Landmarks.},
  author={Pozanco, Alberto and Yolanda, E and Fern{\'a}ndez, Susana and Borrajo, Daniel and others},
  booktitle={IJCAI},
  pages={4808--4814},
  year={2018}
}

@inproceedings{gragera2025gains,
  title={On the gains from using action observations in domain repair},
  author={Gragera, Alba and Fuentetaja, Raquel and Olaya, {\'A}ngel Garc{\'\i}a and Fern{\'a}ndez, Fernando},
  booktitle={Proceedings of the International Conference on Automated Planning and Scheduling},
  volume={35},
  number={1},
  pages={343--347},
  year={2025}
}

@inproceedings{morales2024learning,
author = {Morales, Marianela and Pozanco, Alberto and Canonaco, Giuseppe and Gopalakrishnan, Sriram and Borrajo, Daniel and Veloso, Manuela},
year = {2025},
month = {10},
pages = {},
booktitle = {European Conference on Artificial Intelligence (ECAI)},
title = {On Learning Action Costs from Input Plans},
isbn = {9781643686318},
doi = {10.3233/FAIA250976}
}

@article{pozanco2026planning,
  title={Planning Task Shielding: Detecting and Repairing Flaws in Planning Tasks through Turning them Unsolvable},
  author={Pozanco, Alberto and Morales, Marianela and Totis, Pietro and Borrajo, Daniel},
  journal={arXiv preprint arXiv:2604.07042},
  year={2026}
}

\end{document}